\documentclass{edm_article}

\usepackage{subcaption}
\usepackage{enumitem}
\usepackage{caption}
\usepackage{todonotes}
\usepackage{graphicx}
\usepackage{url}
\usepackage{xcolor}

\begin{document}

\title{Pattern-based Knowledge Component Extraction from Student Code Using Representation Learning}

\numberofauthors{7}

\author{
\alignauthor
Muntasir Hoq\\
       \affaddr{NC State University}\\
       \email{mhoq@ncsu.edu}    
\alignauthor
Griffin Pitts\\
       \affaddr{NC State University}\\
       \email{wgpitts@ncsu.edu}
\alignauthor
Tirth Bhatt\\
       \affaddr{NC State University}\\
       \email{tjbhatt@ncsu.edu}
\and
\alignauthor
Aum Pandya\\
       \affaddr{NC State University}\\
       \email{apandya4@ncsu.edu}
\alignauthor
Andrew Lan\\
       \affaddr{University of Massachusetts Amherst}\\
       \email{andrewlan@cs.umass.edu}
\alignauthor
Peter Brusilovsky\\
       \affaddr{University of Pittsburgh}\\
       \email{peterb@pitt.edu}
\and
\alignauthor
Bita Akram\\
       \affaddr{NC State University}\\
       \email{bakram@ncsu.edu}
}

\maketitle

\begin{abstract}
Personalized instruction aims to provide learners with support that adapts to their individual knowledge and progress toward learning objectives. Discovering and tracing Knowledge Components (KCs) is an important step in building accurate models of student learning. However, KC discovery in computer science education is challenging due to the open-ended nature of programming, wide variability in student solutions, and intertwined use of programming structures in code. We address these challenges with a \textit{pattern-based KC} discovery method that uses a data-driven approach to define KCs as recurring structural patterns in student code that reveal persistent patterns of struggle and mastery in students' solutions. We then evaluate the discovered KCs using expert evaluation and statistical student modeling to demonstrate their effectiveness in capturing student learning and struggles. We propose a framework for modeling students' learning by deriving pattern-based KCs from student code through a three-stage process. First, an attention-based code representation model identifies Abstract Syntax Tree subtrees most relevant to code correctness. Second, a Variational Autoencoder abstracts these subtrees into a smooth latent space, capturing structural similarity across student submissions. Third, the resulting representations are clustered into pattern-based KCs. To assess the effectiveness of pattern-based KCs for modeling students' learning, we adapt the Deep Knowledge Tracing model to incorporate these KCs, demonstrating significant improvements in predictive performance over baseline KT methods. Additionally, the learning curve analysis showed alignment between the derived KCs and learning theory.
\end{abstract}


\section{Introduction}

Student modeling provides the basis for which adaptive educational systems can track knowledge acquisition, adapt instructional strategies, and provide personalized, targeted support~\cite{shi2023kc}. A central part of student modeling is discovering Knowledge Components (KCs) that students practice during problem-solving. Here, KCs refer to discrete units of acquired knowledge or skill inferred from performance on a set of related tasks~\cite{koedinger2012knowledge}.

Modeling learning at the KC level involves two steps: task modeling and evidence modeling~\cite{arieli2019expanded}. An important aspect of task modeling involves specifying the target KCs associated with a given set of learning activities (KC discovery)~\cite{shi2023kc}. Prior work has approached KC discovery through expert-driven design~\cite{alencar2025integrating}, neural approaches that jointly learn latent abstract KCs and their associations with problems from student code while enforcing learning-curve–based constraints~\cite{shi2023kc}, and, more recently, Large Language Models (LLMs) that infer candidate KCs directly from instructional content or student work \cite{tan2024automated,moore2024automated,duan2025automated}. In evidence modeling, traces of students' problem-solving behavior should be mapped to the discovered KCs to enable effective tracing of their learning progress (KC attribution)~\cite{shi2024evaluating}. The accuracy of this mapping directly influences the effectiveness of knowledge tracing and learning analytics, and consequently determines how well student knowledge can be modeled and personalized instruction can be delivered.


In domains such as mathematics and chemistry, problem-solving paths are relatively constrained, and KC granularity is often predefined~\cite{tan2024automated}. In computer science (CS) education, however, KC discovery and attribution pose significant challenges due to the context-dependent nature of what constitutes a unit of knowledge, the vast state space of possible solutions to the same problem, the many programming structures that can represent the same KC, and the interconnected nature of programming patterns \cite{alencar2025integrating}. First, there is no agreed-upon ground truth list of KCs for introductory programming courses \cite{shi2023kc}. Second, the open-ended nature of programming allows for numerous correct implementations; students may solve the same problem using very different structures and strategies, such as recursion vs. iteration or modular decomposition vs. inline logic~\cite{rivers2017data}. Third, even when students use the same programming construct (e.g., a conditional or loop), that construct can serve different conceptual roles depending on context. such as boundary checking, control flow, or decision-making, making it difficult to reliably attribute a single KC. Fourth, the granularity of KCs in programming remains undefined; while low-level constructs like loops and conditionals can be considered fine-grained KCs, high-level KCs often emerge from combinations of these primitives, reflecting more complex problem-solving strategies or conceptual understanding, and can become relevant more over the course of time~\cite{huang2023supporting}. Finally, substantial variation in code structure and expression style across student submissions introduces noise into KC attribution, complicating standardized mappings between code and underlying knowledge components. These challenges render expert-driven KC discovery, common in cognitive tutors and intelligent tutoring systems for many fields, and manual KC allocation infeasible for programming education \cite{rivers2016learning}. 

In response, recent approaches have sought to discover KCs as constructs that students commonly acquire during problem-solving, as reflected in their code submissions over time. Prior work has leveraged code parsers~\cite {rivers2016learning,alencar2025integrating}, deep learning models~\cite{shi2024knowledge,shi2023kc}, as well as LLMs~\cite {niousha2025llm,duan2025automated} to infer KCs from student code. However, these methods face limitations, including a lack of explainability and limited granularity of discovered KCs, often focusing on programming constructs such as loops or conditionals rather than capturing more nuanced conceptual understanding.

To address these challenges, we propose a framework for automated KC discovery from student programming submissions. Unlike traditional KCs, which are often defined in terms of skill-tags or curriculum objectives, our approach defines KCs as structural patterns, which we introduce as \textit{pattern-based KCs}. These pattern-based KCs are recurring and semantically meaningful subtree patterns extracted from the Abstract Syntax Trees (ASTs) of student code, reflecting programming patterns and algorithmic constructs necessary for correctly solving a programming problem. 

Our framework begins with the Subtree-based Attention Neural Network (SANN)~\cite{hoq2023sann}, trained to predict code correctness while assigning attention weights to relevant AST subtrees~\cite{hoq2025automated}. These subtrees are then normalized to remove literal and identifier-specific variance, encoded into a latent space using a Variational Autoencoder (VAE) to generate important patterns from the code and abstract them in a smooth latent space to capture similarities across student submissions. These patterns are then clustered using K-means to form the discovered pattern-based KCs. 

To demonstrate how our KC structures can support modeling of student learning, we adapt the Deep Knowledge Tracing (DKT) model~\cite{piech2015deep} to incorporate discovered pattern-based KCs. Experimental results show that our approach improves the predictions of student performance when incorporated in the traditional DKT method and produces meaningful learning trajectories, providing a foundation for detecting mastery and guiding instructional decision-making. To further examine the alignment of pattern-based KCs with learning theory, we analyze student learning curves to track how mastery develops across different KCs~\cite{rivers2016learning}. Finally, we validate the educational relevance and internal coherence of the discovered KCs through expert evaluation, and illustrate their explainability with a case study tracing a representative KC across multiple problems and student solutions.

The main contributions of this paper are fourfold. First, we formalize \textit{pattern-based KCs} as a new conceptual representation of knowledge in CS education, grounded in recurring and semantically meaningful code structures. Second, we develop a scalable and explainable framework for automatically discovering these KCs from student code submissions using attention-based modeling and representation learning. Third, we demonstrate that integrating pattern-based KCs into a knowledge tracing model significantly improves predictive accuracy for student performance modeling compared to traditional approaches. Fourth, we provide empirical validation of the educational utility and coherence of the discovered KCs through learning curve analysis, expert evaluation, and explainability case studies. These contributions advance explainable knowledge modeling in programming education, enabling more reliable tracking of how students acquire programming skills and providing actionable insights for researchers and educators seeking to support individualized instruction.

\section{Related Work}
Understanding how learners acquire and apply specific skills is a central goal of educational research, and Knowledge Components (KCs) provide a structured way to represent this learning. We review prior work relating to the conceptualization of KCs and existing methods for their automated discovery, with a focus on programming education.

\subsection{Knowledge Components}
The concept of Knowledge Components (KCs) is foundational in educational research, particularly within cognitive science and intelligent tutoring systems. KCs are defined as discrete units of knowledge or skills that a learner must acquire to perform a task successfully. They encompass a range of cognitive elements, including concepts, principles, facts, and procedures~\cite{koedinger2012knowledge,rivers2016learning}.

The Knowledge-Learning-Instruction Framework~\cite{koedinger2012knowledge} further elaborates on KCs by linking them to instructional events and learning outcomes. According to this framework, KCs are the mental constructs that mediate between instructional activities and observable learning behaviors. This perspective underscores the importance of accurately discovering and modeling KCs to enhance instructional design, learning analytics, and develop Intelligent Tutoring Systems \cite{corbett1997intelligent}. In programming education, defining KCs and determining their appropriate level of granularity is inherently challenging, as they can span from fine-grained concepts, e.g., loop initialization, to complex interactions among constructs such as nested conditionals within loops, encompassing control structures, data manipulation, and algorithmic strategies. Accurately modeling these components is crucial for developing adaptive learning systems that can provide insights into student knowledge tracing~\cite{liang2022help}, help provide personalized feedback, recommend worked examples, and support to learners~\cite{abdelrahman2023knowledge}.

\subsection{Knowledge Component Discovery}
Automated discovery of KCs from student code has gained attention, aiming to mitigate the labor-intensive nature of manual annotation and to scale personalized education.

Deep learning approaches have been used to model student learning and discover KCs by analyzing code submissions. These AST-based methods leverage the structural representation of code to assign KCs~\cite{shi2024knowledge,filgueira2022inspect4py,rivers2016learning}. For instance, the KC-Finder model integrated learning curve theory into its architecture to discover KCs that align with students' learning progression~\cite{shi2023kc}. Although these models can assign KCs, they often suffer from explainability issues, making it challenging to understand the rationale behind KC assignments. Moreover, these approaches may struggle with the semantic understanding of code, especially when different code structures achieve the same functionality, and understanding the granularity and hierarchical nature of different KCs could be challenging.

LLMs have recently been explored for KC discovery due to their ability to understand programming code and generate human-like text. Studies have demonstrated that LLMs can discover KCs from code and problem descriptions~\cite{niousha2025llm,niousha2024use,duan2025automated,fan2025adaptive}. Despite their potential, LLMs can produce overgeneralized KCs, i.e., different conditional structures can not be identified effectively due to a lack of KC taxonomy, or even irrelevant KCs. While manual KC tagging by experts might ensure high accuracy, it is impractical for large-scale applications due to the significant time and effort required~\cite{shi2024knowledge}. Moreover, inconsistencies among experts can lead to variability in KC definitions, affecting the reliability of the annotations. While automated methods for KC discovery have advanced, challenges remain in balancing accuracy, explainability, and scalability. Our work aims to address these issues by introducing a pattern-based approach that leverages attention mechanisms and clustering techniques to discover meaningful KCs from student code submissions.

\section{Dataset}
We conduct our work using a publicly available dataset\footnote{\url{https://pslcdatashop.web.cmu.edu/Files?datasetId=3458}} obtained from the CodeWorkout platform~\cite{edwards2017codeworkout}. The dataset contains Java programming submissions collected during an introductory CS course offered at a US public university in Spring 2019. In total, the dataset spans $50$ programming problems distributed across $5$ course assignments, along with students’ normalized final exam scores (scaled between $0$ and $1$). No demographic or personally identifiable information is included in the dataset. Each submission is evaluated based on the number of passed test cases and is assigned a correctness score between $0$ and $1$. For the purpose of this work, we binarize these labels into correct (all test cases passed) and incorrect (one or more test cases failed) classes. 
The dataset comprised a total of $47{,}764$ code submissions from $407$ students, of which $18{,}787$ were correct submissions and $28{,}977$ were incorrect. 

\section{Methodology}
We aim to model students' learning progress by discovering KCs as structural patterns in code. We refer to these KCs as \textit{pattern-based KCs}, representing actionable and explainable units that capture students' focal programming knowledge and skills. To discover pattern-based KCs from student code submissions, our approach combines (i) fine‐grained signals that pinpoint the code segments most critical for program correctness and (ii) context‐aware abstraction that groups structurally distinct yet semantically related implementations under the same conceptual pattern. First, we \emph{extract relevant structural patterns from student programs} using SANN~\cite{hoq2025automated}, an attention-based model designed for program analysis, which applies hierarchical AST processing and attention scores to reveal which subtrees drive predictions of program correctness. However, SANN is trained to separate programs by correctness and does not strictly learn a continuous latent space where structurally similar subtrees can be clustered together. To address this, we \emph{generate context-aware representations of influential patterns} using a VAE. By reconstructing sequences of high‐attention subtrees, the VAE (i)~injects broader context, capturing how salient patterns interact within the complete program and across submissions, and (ii)~regularizes them into a continuous latent manifold that preserves embedding similarity. 

This latent representation provides the basis for \emph{clustering related patterns into distinct pattern-based KCs}. To discover these KCs, we train the VAE to generate representative patterns corresponding to focal programming concepts and skills present in the input AST. Due to the randomness involved in generating programming patterns, multiple patterns might represent the same focal concept and skill. Thus, we apply K-means clustering to group similar patterns into clusters that represent distinct pattern-based KCs. K-means is suitable here since its Euclidean distance assumption aligns well with the smooth, isotropic structure of the VAE model's learned latent manifold. These clusters form the set of pattern-based KCs used in our framework. Finally, we \emph{advance} student modeling by integrating the discovered pattern-based KCs into a Deep Knowledge Tracing model~\cite{piech2015deep}, enabling more accurate prediction of student performance. Furthermore, to assess educational relevance, we perform learning curve analysis to test whether the KCs align with established learning theory~\cite{rivers2016learning}.

\subsection{Extracting Relevant Structural Patterns\\ from Student Programs}
The first stage of the framework focuses on pinpointing the structural elements in student code that are most critical for program correctness. These elements correspond to programming patterns required to meet problem requirements and are often sources of student difficulty. To extract them, we leverage a modified Subtree-based Attention Neural Network (SANN) \cite{hoq2025automated}, an explainable architecture that learns from the hierarchical structure of code and identifies logical errors in incorrect implementations. SANN has been previously validated across different tasks, including program algorithm classification, detecting LLM-generated code, and recommending relevant worked examples~\cite{hoq2023sann,hoq2024detecting,hoq2025worked}.

\begin{figure}[h]
  \centering
  \includegraphics[width=\columnwidth]{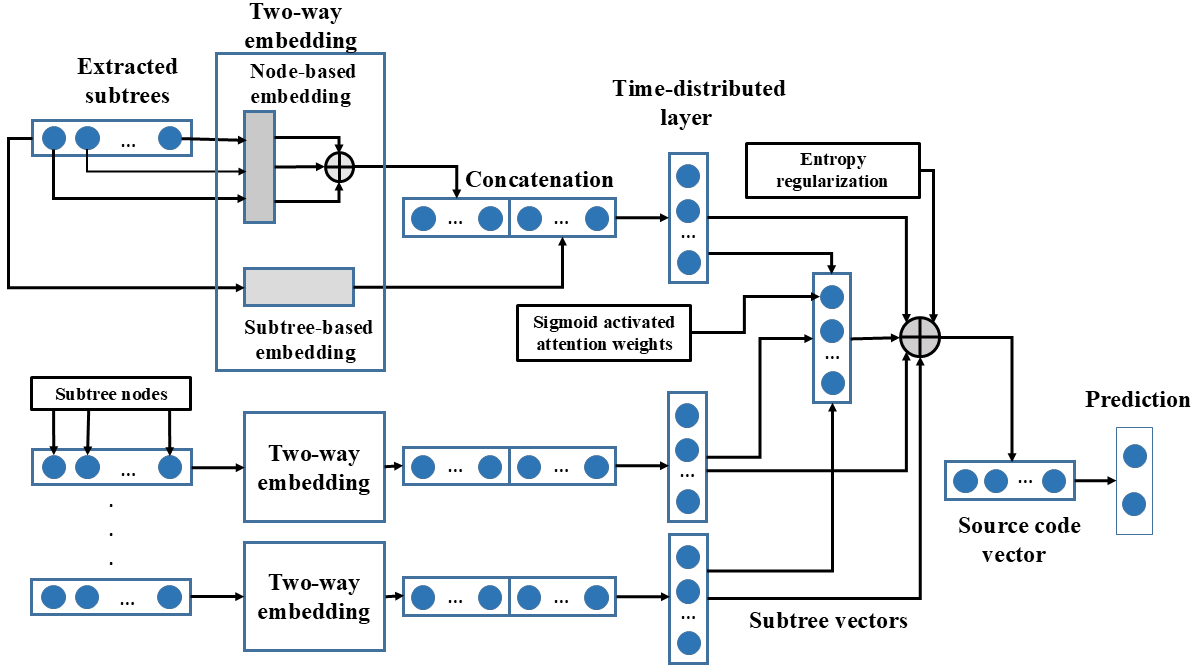}
  \caption{Architecture of the SANN model.}
  \label{fig:sann}
  \Description{Architecture of the SANN model with the two-way embedding approach and the attention neural network.}
\end{figure}

SANN operates on code ASTs by extracting subtrees at multiple depths, which may overlap due to the hierarchical structure of programs. This overlap allows the model to capture coarse-grained constructs (e.g., full conditionals) and fine-grained elements (e.g., comparison expressions) that can independently contribute to correctness~\cite{hoq2024towards}. Trained to classify student submissions as correct or incorrect, the model produces attention weights indicating which subtrees most influence predictions (Figure~\ref{fig:sann}). In correct submissions, high-attention subtrees typically represent programming patterns necessary for solving the problem correctly, such as loops, conditionals, or combinations of multiple concepts. In incorrect submissions, high-attended subtrees often reflect logical errors or misconceptions, providing insight into why the submission failed~\cite{hoq2025automated}.

Within our framework, these high-attention subtrees form the initial candidates for pattern-based KCs. They are carried forward into the next stage, where the VAE abstracts away surface-level variation and organizes them into a latent space suitable for clustering into representative KCs.

\subsection{Latent Representation of Candidate KCs}

The second stage of the framework uses a $\beta$-VAE model ~\cite{higgins2017beta} to generate latent representations of the most influential programming patterns inferred from student code solutions. These representations form the basis for modeling students' learning over time.

\begin{figure}[h]
  \centering
  \includegraphics[width=0.95\columnwidth]{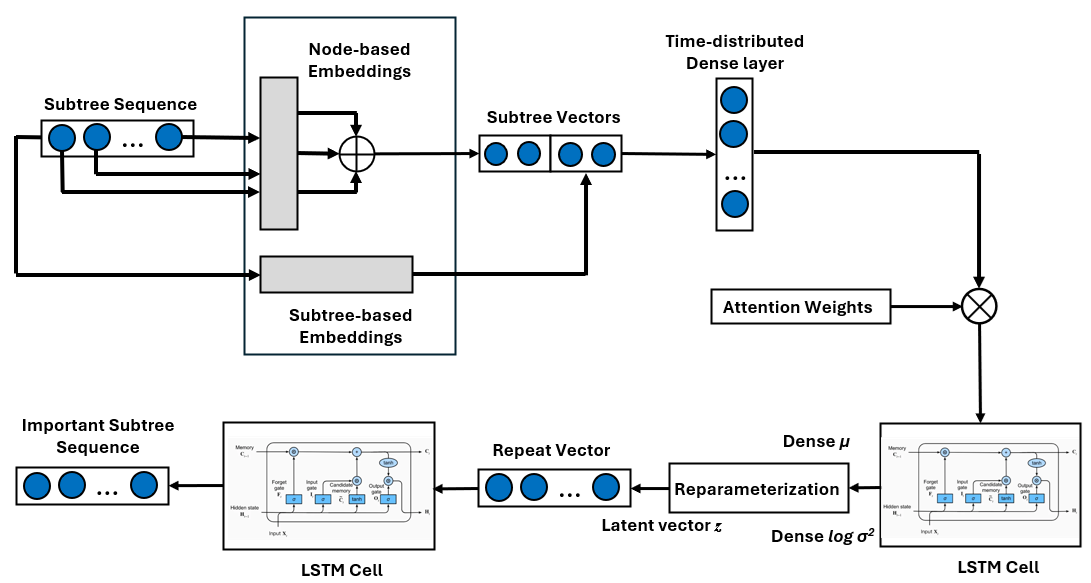}
  \caption{Architecture of the $\beta$-VAE model.}
  \label{fig:vae}
  \Description{Architecture of the VAE model.}
\end{figure}

As a first step, we abstract away surface-level syntactic differences, such as specific variable names, literal values, and function identifiers, while preserving the underlying semantic structure. Although such details can be useful for correctness classification, they introduce redundant variation when training the $\beta$-VAE and add noise to KC discovery ~\cite{shi2023kc}. This abstraction ensures the VAE focuses on the structural and semantic features of code rather than superficial syntactic differences. The normalized ASTs and their associated high-attention subtrees are then used to train the $\beta$-VAE model. During training, subtrees are encoded into continuous latent representations and reconstructed, with the model designed so that semantically similar subtrees occupy neighboring regions in latent space. This property enables grouping identified patterns into representative pattern-based KCs. A schematic of the $\beta$-VAE architecture is shown in Figure~\ref{fig:vae}.

To encode subtrees, the VAE model embeds each along two dimensions: the sequence of AST nodes within the subtree and the subtree itself. Node sequences are passed through a time-distributed embedding layer with positional encoding to preserve the relative ordering of nodes within each subtree. This is essential for capturing nuanced structural differences, especially when similar logic is implemented with different nesting or sequencing. The embedded node sequences are then fed into a time-distributed LSTM layer, which compresses each node-level sequence into a fixed-length vector representation.

In parallel, subtree identifiers are embedded and concatenated with the output of the node-level LSTM to form composite embeddings. These combined vectors are then processed by a time-distributed dense layer with a non-linear activation to produce the final subtree representations. To emphasize the most relevant patterns, we scale each vector by the attention weights from SANN, giving greater influence to subtrees most strongly linked to correctness.

The weighted sequence of subtree vectors is passed through an LSTM to capture their sequential order, with dropout applied for regularization. The output is then mapped to two vectors representing the mean, \( \boldsymbol{\mu} \), and log-variance, \( \log \boldsymbol{\sigma}^2 \), of the latent distribution. From these, we sample a latent vector, \( \mathbf{z} \), using reparameterization from the distribution:
\[
\mathbf{z} = \boldsymbol{\mu} + \boldsymbol{\sigma} \odot \boldsymbol{\epsilon}, \quad \text{where } \boldsymbol{\epsilon} \sim \mathcal{N}(0, I).
\]
To ensure the latent space remains smooth and structured, the KL divergence loss is added, 
encouraging the distribution to stay close to a standard Gaussian:
\[
\mathcal{L}_{\text{KL}} = -\frac{1}{2} \sum_{j=1}^{d} \left(1 + \log \sigma_j^2 - \mu_j^2 - \sigma_j^2 \right).
\]
The decoder reconstructs the original subtree sequence by repeating the latent vector across the sequence dimension and passing it through an LSTM, followed by a dense output layer with softmax activation. Reconstruction loss is calculated using categorical cross-entropy between the predicted and true subtree tokens:
\[
\mathcal{L}_{\text{recon}} = - \sum_{t=1}^{T} \log p(\hat{y}_t \mid \mathbf{z}).
\]
The total VAE loss is a weighted combination of reconstruction and KL divergence losses, with a tunable parameter \( \beta \) controlling the trade-off:
\[
\mathcal{L}_{\text{total}} = \mathcal{L}_{\text{recon}} + \beta \cdot \mathcal{L}_{\text{KL}}.
\]
Within the framework, the trained VAE builds on SANN by reconstructing the sequences of high-attention subtrees, capturing broader program context that SANN alone cannot. In doing so, it acts as a denoising compressor, generating representative patterns of important subtrees from correct submissions that preserve the essential aspects of program structure. This design also enables the analysis of incorrect submissions: during inference, sequences of subtrees from incorrect code are passed through the model, and those with high attention weights, identified by SANN, are examined as sources of logical errors~\cite{hoq2025automated}. The VAE can then suggest corrected structural equivalents for these faulty segments by leveraging both the local context of the faulty subtrees and the global structure of the code. This mapping facilitates the construction of learning curves based on KC mastery, which in turn enables the evaluation of KC models~\cite{rivers2016learning}. In this way, the VAE produces structural KC representations that generalize across variation while preserving conceptual integrity, providing the foundation for defining \emph{pattern-based KCs}. The next step is to consolidate these representations into stable and explainable units of knowledge.

\subsection{Clustering Pattern-based KCs}

The third stage consolidates the previously generated latent representations into representative pattern-based KCs. These KCs serve as abstractions of the minimal functional substructures required to solve a programming problem, independent of surface-level syntactic differences. Clustering is performed using the encoded subtree vectors from the VAE encoder, which embed normalized high-attention subtrees drawn from correct student submissions. Before clustering, we filter out anomalous student behaviors that could distort KC interpretation. Consistent with prior work~\cite{hoq2024explaining,shi2023kc}, we observe that some students exhibit suspicious learning patterns, such as frequent failures on simple problems followed by sudden, consistent success on more advanced problems without intermediate improvement. These cases, which may indicate cheating or code sharing, can skew the KC distribution and are therefore excluded to preserve the integrity of the discovery process.

After filtering, we cluster subtree vectors from correct code submissions, which are expected to reflect structurally sound and semantically valid constructs. K-means clustering is used to group subtrees based on their similarity in the learned latent space, under the assumption that subtrees with nearby encodings represent similar underlying knowledge. 
The appropriate number of clusters is determined via the elbow method, a commonly used heuristic that identifies the point of diminishing returns in the variance explained by additional clusters.

Each cluster defines a pattern-based KC, represented by its centroid as the canonical example of that KC. High-attention subtrees from both correct and incorrect submissions are then assigned to their nearest centroid, linking each to a representative KC. This enables our proposed framework, which provides a consistent and explainable way to track student learning through code submissions. The presence or incorrect student implementation of a pattern-based KC can be measured across attempts, and the underlying code pattern responsible for the resultant pattern-based KC can be traced back through the pipeline. In doing so, the framework scales automated knowledge modeling in programming education while remaining grounded in structurally meaningful code representations.

\subsection{Improving Deep Knowledge Tracing with Pattern-based KCs}

We incorporate the discovered pattern-based KCs into the original DKT framework~\cite{piech2015deep} to advance student modeling by providing a more precise representation of learners’ knowledge. In DKT, student mastery is modeled as a sequence of interaction events, with the model estimating the probability of success on future tasks. We hypothesize that incorporating more precise KC-level information will improve the performance of DKT and thereby support more accurate tracing of students’ learning progress.

Formally, let \( S = \{x_1, x_2, \dots, x_T\} \) represent a sequence of \( T \) student interactions. Each interaction \( x_t \) corresponds to the student's attempt on a problem at time \( t \), consisting of a problem identifier \( q_t \in \{1, 2, \dots, M\} \) and a correctness label \( a_t \in \{0,1\} \), where \( M \) is the total number of unique problems. Following the original DKT formulation, each interaction is encoded as a one-hot vector of size \( 2M \). Specifically, the element at index \( q_t + M \cdot (1 - a_t) \) is set to 1, indicating whether the student answered the problem correctly or incorrectly, and all other elements are set to 0.

In our enhanced model (KC-DKT), the input is extended by incorporating the pattern-based KC mastery vector associated with each submission. This process results in a binary vector of length \( K \), where \( K \) is the number of pattern-based KCs. For each submission, this vector encodes the presence (1) or absence/incorrect implementation (0) of each pattern-based KC in the student’s solution. The input \( \tilde{x}_t \) to the DKT model at time \( t \) is constructed by concatenating the original DKT problem-correctness encoding with the KC vector:
\[
\tilde{x}_t = [x_t^{\text{DKT}} \parallel x_t^{\text{KC}}].
\]

\begin{figure}[t]
  \centering
  \includegraphics[width=0.95\columnwidth]{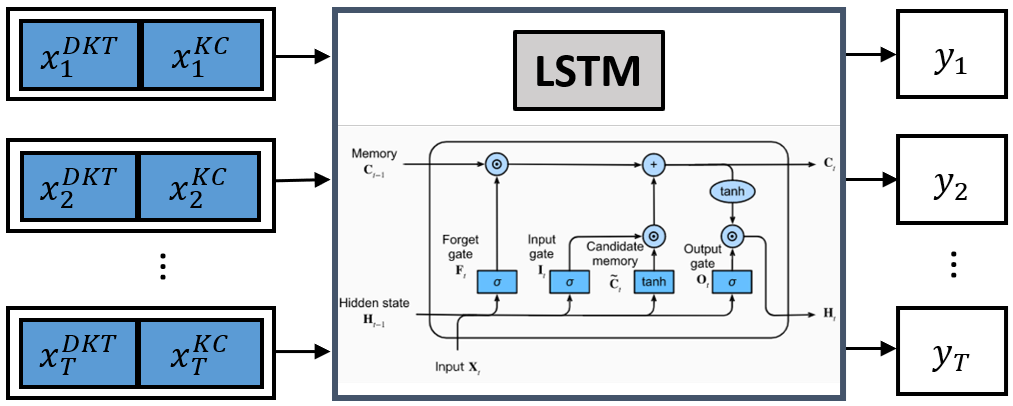}
  \caption{KC-DKT model with pattern-based KC vectors.}
  \label{fig:dkt}
  \Description{Architecture of the improved DKT model.}
\end{figure}

The KC-DKT architecture (see Figure~\ref{fig:dkt}), implemented as an LSTM, receives the input sequence \( \{\tilde{x}_1, \tilde{x}_2, \dots, \tilde{x}_T\} \) and outputs predictions \( \{y_1, y_2, \dots, y_T\} \), where each \( y_t \in \mathbb{R}^M \) gives the probability of successful implementation on each of the \( M \) problems. The predicted probability for the next attempted problem \( q_{t+1} \) is obtained from the corresponding entry in \( y_t \). The model is trained using binary cross-entropy loss and evaluated using the Area Under the Receiver Operating Characteristic Curve (AUC) metric to assess performance in distinguishing correct versus incorrect submissions.


\subsection{Educational Utility of Pattern-based KCs}
Along with improving predictive knowledge tracing, we assess whether the discovered pattern-based KCs are educationally meaningful and suitable for modeling student learning. We evaluate this along three dimensions: learning curve analysis to check alignment with learning theory, an explainability case study, and pedagogical coherence through expert evaluation of KC clusters.

\subsubsection{Learning Curve Analysis}
Learning curve analysis is a standard approach for evaluating whether inferred knowledge components align with the learning trajectory of student knowledge over time. Following conventions in programming education research~\cite{rivers2016learning,shi2023kc,alencar2025integrating}, we define an \textit{opportunity} as a student's first attempt on a problem that exercises a given KC. Restricting to first attempts reduces the confounding effects of feedback or guessing in later submissions. For each pattern-based KC, we compute the average error rates across increasing opportunity counts and analyze the resulting learning curves. KCs that display a downward error trend with increasing practice are categorized as \textit{good learning}, while flat or erratic curves suggest \textit{still-learning} or \textit{no-learning}~\cite{rivers2016learning,shi2023kc}. These profiles indicate whether the discovered KCs align with the learning theory and provide a basis for instructional interpretation.


\subsubsection{Evaluating Pattern-based KC clusters}
\label{method:educational_utility}

To validate whether the discovered pattern-based KCs are pedagogically meaningful and internally coherent, we conducted a structured expert evaluation of the KC clusters. For each cluster, a representative medoid pattern was selected (the subtree closest to the cluster center in latent space) and sampled ten additional patterns from the same cluster. To ensure robustness, five patterns were selected based on proximity to the medoid and five based on proximity to the cluster centroid using cosine similarity.

Three expert evaluators in programming education assessed each cluster. Evaluators first judged whether the medoid represents an important programming pattern that students should learn in introductory programming courses (educational utility). They then compared each candidate pattern to the medoid, judging whether it reflected the same underlying programming idea (conceptual consistency) and how similar the pattern’s operation on program state or control flow was to that of the medoid (action-level similarity) on a scale of 1 to 4 (1: Completely different/unrelated action, 2: Related action, with weak relationship, 3: Same action type, minor variation, and 4: Same action type and same role). Evaluators were instructed to focus on conceptual intent and control structure (e.g., branching, iteration, accumulation), while ignoring superficial differences such as variable names, formatting, or literal values.

To establish reliability, all evaluators independently annotated an initial $10$\% of the clusters. Inter-rater agreement on this subset was perfect for judgments of educational utility and conceptual consistency, and exceeded $0.92$ for action-level similarity. After resolving minor disagreements and aligning interpretations, the remaining clusters were divided among evaluators and annotated independently. This protocol provides an external validation of both the educational relevance and coherence of the pattern-based KCs.

\subsubsection{Explainability of Pattern-based KCs}
To examine the explainability of the discovered pattern-based KCs, we conducted a case study tracing a representative KC across multiple problems and student submissions. Using the projection of attention-weighted subtrees back onto source code, we annotated the portions of students' code associated with the KC in both correct and incorrect attempts. This procedure enabled us to follow how a single KC manifested in successive student submissions and across different problems, providing a basis to evaluate whether the KCs capture consistent and reusable conceptual patterns.

\section{Experiments}

We evaluated the proposed framework by training each component in sequence and examining its ability to discover meaningful pattern-based KCs from student code. The experiments proceeded in three stages: training the modified SANN model to identify code substructures most relevant to program correctness, learning a latent representation of these subtrees with a $\beta$-VAE, and clustering the resulting embeddings into representative KCs. The resulting set of KCs was then used for downstream analyses, including integration into a KC-enhanced DKT framework and learning curve evaluation. On average, each programming problem was associated with about 24 KCs (23.68, SD = 6.02), each student submission involved roughly 4 KCs (3.87, SD = 1.56), and each KC was observed across approximately 25 problems (25.36, SD = 17.28), indicating broad coverage of programming constructs throughout the dataset.



In the first stage, we trained the modified SANN model to classify Java submissions as correct or incorrect while assigning attention weights to AST subtrees. The full dataset, consisting of $18{,}787$ correct and $28{,}977$ incorrect submissions across $50$ problems, was split into training, validation, and test sets with an 80:10:10 stratified split to preserve label proportions. The model was trained using a code embedding size of $128$ with a maximum of $100$ nodes per subtree and $100$ subtrees per program, for up to $100$ epochs with an early stopping patience of $20$. We used the Adamax optimizer with a learning rate of $0.001$ and minimum entropy regularization weight of $3 \times 10^{-5}$ to guide the attention mechanism without compromising classification performance. The trained model achieved high scores across all evaluation metrics: accuracy ($0.87$), precision ($0.86$), recall ($0.88$), and F1-score ($0.87$), confirming its ability to capture correctness-related structural patterns. Attention weights and associated subtrees were extracted for the next stage.

In the second stage, we filtered subtrees with attention weights above $20\%$, following the original study of this model~\cite{hoq2025automated}. From this filtering, we obtained a total of $78,647$ important subtrees from correct submissions and $132,658$ important subtrees from incorrect submissions. Identifier tokens were normalized to reduce surface variation, and a $\beta$-VAE was trained exclusively on high-attention subtrees from correct code to ensure that the latent space reflected valid and pedagogically meaningful structures. The VAE used a subtree embedding size of $64$, latent dimension $128$, intermediate dense layers matching the embedding size, dropout rate $0.2$ (to prevent overfitting), and was trained for up to $30$ epochs (patience $10$) with Adamax optimizer ($\text{lr}=0.001$) and $\beta=1\times 10^{-2}$ to balance reconstruction fidelity with latent space regularization. The VAE achieved $95\%$ reconstruction accuracy across all tokens, though only $10\%$ when considering important subtrees in isolation. This lower score was expected, since the objective was not perfect reconstruction but learning a latent embedding space where structurally similar subtrees are mapped to nearby regions~\cite{higgins2017beta}. In practice, the VAE compressed $11,953$ unique subtree patterns into $106$ unique representative outputs, demonstrating its ability to compress and abstract fine-grained structural variations. To validate semantic alignment, we compared cosine similarity scores between predicted and input subtrees using CodeT5 (pretrained programming language model, 220 million parameters) embeddings~\cite{wang2021codet5}, against a randomized baseline. Our proposed approach achieved a significantly higher mean similarity score of 0.96 (standard deviation 0.03) compared to the randomized baseline's mean similarity of 0.85 (standard deviation 0.04) between the input and predicted representative subtrees. A statistical test confirmed that the difference between the two distributions was significant with a $p$-$value$ of $0.0001$.

In the final stage, we clustered the VAE-generated subtree vectors into distinct pattern-based KCs using K-means. Only high-attention subtrees from correct submissions were considered, ensuring clusters reflected valid programming patterns. The elbow method indicated an optimal cluster count of $k=50$, balancing granularity with interpretability. Each subtree vector was assigned to its nearest cluster center, with the centroid serving as the representative pattern for that KC. Both correct and incorrect subtrees were then mapped to their closest KC, enabling downstream analyses of student performance at the KC level. This process produced a set of $50$ structurally coherent and pedagogically grounded KCs extracted directly from student code, providing a foundation for the subsequent integration into our KC-DKT model and learning curve evaluation.

\section{Results}
In this section, we evaluate the proposed pattern-based KC framework from three complementary perspectives. First, we examine whether the discovered KCs improve predictive student modeling through knowledge tracing. Second, we assess whether the KCs exhibit learning dynamics consistent with learning theory via learning curve analysis. Finally, we validate the educational relevance and internal coherence of the KC clusters through an expert evaluation.

\subsection{Deep Knowledge Tracing}

To demonstrate that our discovered pattern-based KCs improve on prior approaches for modeling and predicting student learning over time, we integrated them into the original DKT model~\cite{piech2015deep}. Unlike previous approaches that used raw code vectors as input to the DKT model~\cite{hoq2025automated,yang2022code}, our \emph{KC-DKT} model incorporated a mastery vector that encodes, for each of the $50$ discovered pattern-based KCs, whether it was correctly demonstrated in a student’s submission.

Our KC-DKT model was trained to predict whether the student would succeed on the next attempt based on their KC state history. To make the evaluation fair, we split the dataset at the student level using an 80:10:10 ratio for training, validation, and testing, ensuring no overlap between students across splits. Hyperparameter optimization was performed using Keras Tuner. 
The best performing configuration used an LSTM layer with $512$ hidden units, a dropout rate of $0.1$, and a learning rate of $0.01$. Our KC-DKT model using pattern-based KC vectors achieved an AUC of $74.77$\%, outperforming the original DKT baseline~\cite{piech2015deep}, which achieved an AUC of $65.87$\%. Assignment-wise breakdowns of AUC scores are provided in Table~\ref{tab:dkt}, alongside a comparison with Bayesian Knowledge Tracing (BKT)~\cite{badrinath2021pybkt} and Code-DKT~\cite{yang2022code}, which uses AST-derived code vectors for single assignments.

\begin{table}
\centering
\caption{Knowledge tracing AUC (\%) averaged over 10 runs with different random splits. *($p<0.01$), **($p<0.001$) indicate significant improvements of KC-DKT over DKT.}
\scalebox{.92}{
\begin{tabular}{|l|c|c|c|c|c|c|} \hline
\textbf{Model} & \textbf{A1} & \textbf{A2} & \textbf{A3} & \textbf{A4} & \textbf{A5} & \textbf{All} \\ \hline
BKT & 61.6 & 57.5 & 54.1 & 56.3 & 53.2 & 56.5 \\ \hline
DKT & 71.2 & 73.1 & 76.8 & 69.2 & 75.1 & 65.9 \\ \hline
Code-DKT & 74.3 & 76.6 & 80.4 & 72.8 & 79.1 & - \\ \hline
KC-DKT & \textbf{82.4}** & \textbf{77.6}* & \textbf{80.8}* & \textbf{82.0}** & \textbf{85.4}** & \textbf{74.8}* \\ \hline
\end{tabular}}
\label{tab:dkt}
\end{table}

These results highlight the strength of our pattern-based KC representation for knowledge tracing. 
The approximately $13$\% improvement over the original DKT represents a substantial gain relative to previously reported extensions of knowledge tracing models that incorporate code-level representations, which have typically yielded smaller or more incremental improvements in their respective evaluation settings~\cite{yang2022code,hoq2025automated,choi2020ednet}. 
Having demonstrated predictive advantages, we further examine whether the discovered KCs align with learning theory by analyzing student performance trajectories through learning curve analysis.

\subsection{Learning Curve Analysis}

We next evaluated the discovered pattern-based KCs using learning curve analysis to test whether the KCs align with expected trends from learning theory~\cite{shi2023kc,rivers2016learning,fan2025adaptive,alencar2025integrating}. A learning curve reflects improvement over repeated opportunities to practice a KC, typically showing a decline in error rate as mastery increases. Following prior work in knowledge tracing~\cite{selent2016assistments,badrinath2021pybkt}, we considered only the first submission of each student per problem to avoid confounds from debugging or superficial edits. For each KC, we then computed the average error rate across sequential opportunities, where an opportunity was defined as a student’s encounter with code involving that KC.

\begin{figure*}[h]
    \centering
    \begin{subfigure}[b]{0.3\textwidth}
        \centering
        \includegraphics[width=\linewidth]{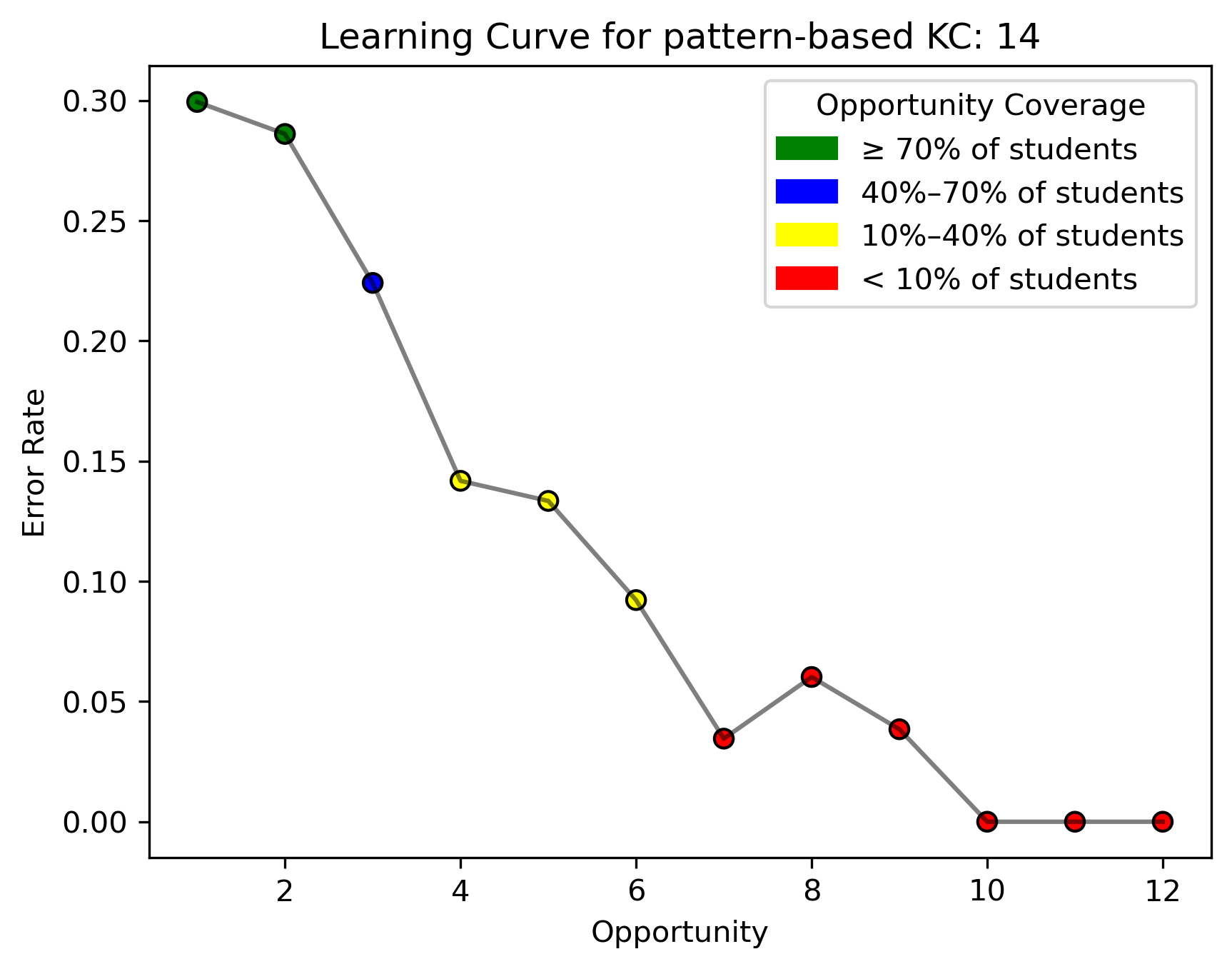}
        \caption{Good learning}
        \label{fig:KC_good}
    \end{subfigure}
    \hfill
    \begin{subfigure}[b]{0.3\textwidth}
        \centering
        \includegraphics[width=\linewidth]{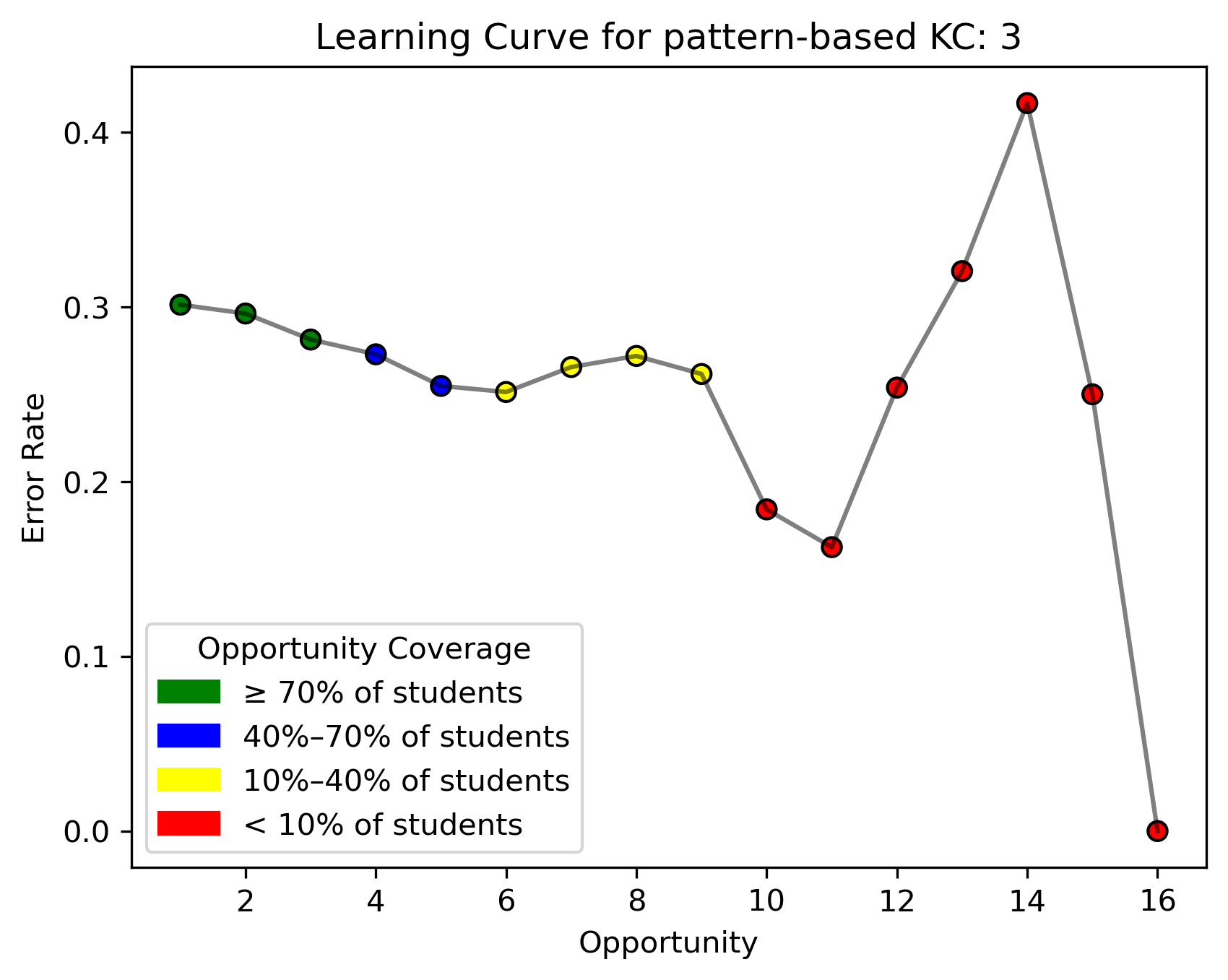}
        \caption{High error / still learning}
        \label{fig:KC_still}
    \end{subfigure}
    \hfill
    \begin{subfigure}[b]{0.3\textwidth}
        \centering
        \includegraphics[width=\linewidth]{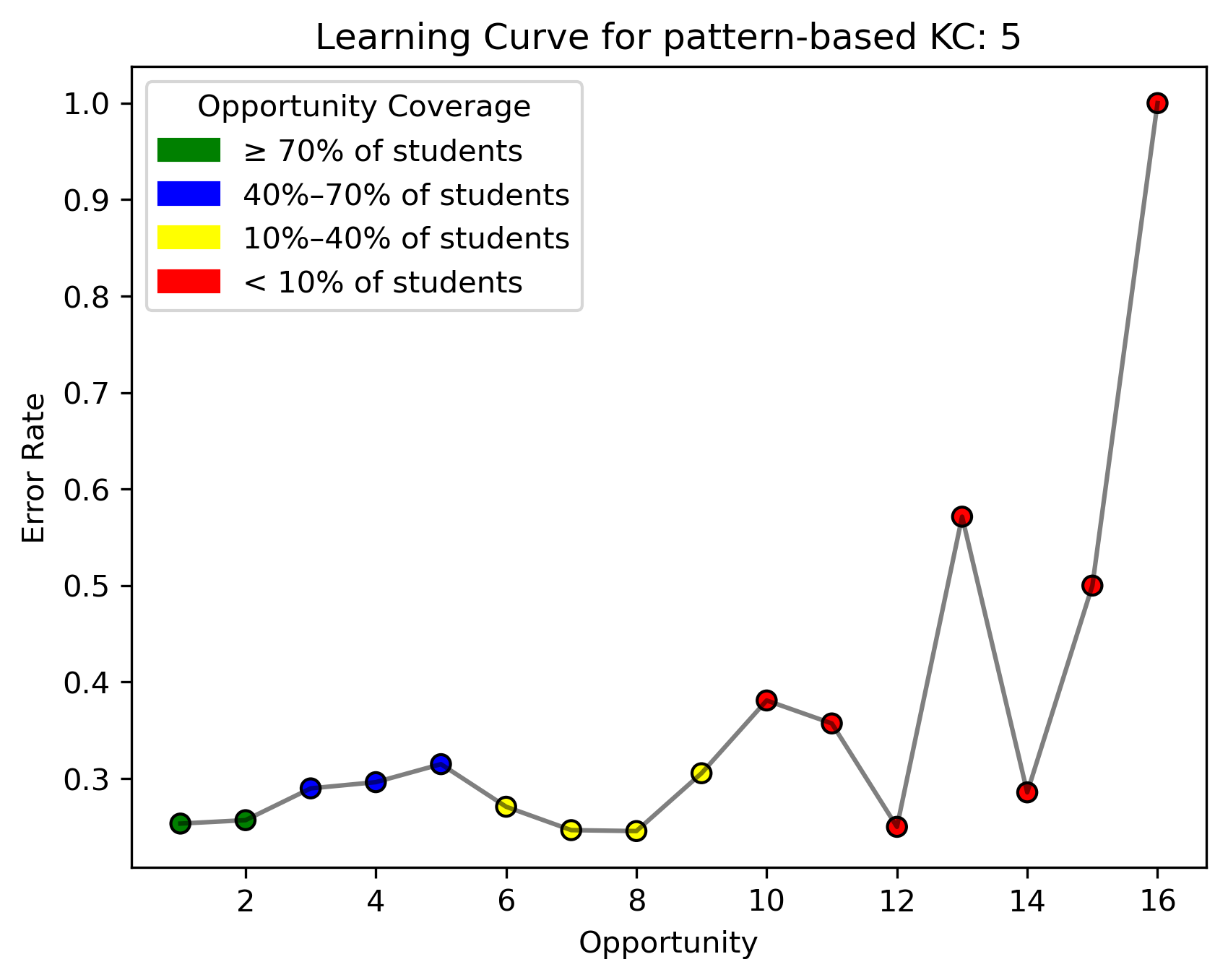}
        \caption{No learning}
        \label{fig:KC_no}
    \end{subfigure}

    \caption{Representative learning curves for 'good learning', 'high error/still learning', and 'no learning' trends. Node
    colors indicate opportunity coverage: the percentage of students who encountered the KC at a given opportunity.}
    \label{fig:three_learning_curves}
\end{figure*}

We observe that the learning curves showed three qualitatively distinct trends~\cite{rivers2016learning,fan2025adaptive} (Figure~\ref{fig:three_learning_curves}). The majority, $35$ out of $50$ pattern-based KCs, showed a steady decline in error rates, which we denote as \textit{good learning}. The learning curves exhibit the expected behavior of a steady decline, indicating that students gradually acquire mastery over these patterns through repeated exposure, with these KCs appearing in an average of 20 problems each (see Figure~\ref{fig:KC_good}).

A second group of clusters, $10$ out of $50$ pattern-based KCs, labeled \textit{high error/still learning}, also trended downwards but showed fluctuations in later opportunities before eventually decreasing (see Figure~\ref{fig:KC_still}). This trend likely reflects the difficulty and recurrence of these patterns across problems. Many of these KCs involve complex conditional structures, such as nested conditionals and complex Boolean expressions, appearing in about $30$ problems per KC.

Finally, $4$ clusters fell into the \textit{no learning} category, where error rates remained high across opportunities (see Figure~\ref{fig:KC_no}). These patterns often combined multiple structural elements or appeared in highly variable set of problems (mean number of problems per KC: $37$). Such "no learning" curves are not unique to our approach but occur in KC modeling more generally, reflecting challenges in reliably measuring genuine learning progress amid noise \cite{rivers2016learning}. Apart from these categories, one additional cluster lacked sufficient opportunities to generate a reliable curve.

    

Overall, the learning curve analysis demonstrates that our pattern-based KCs capture meaningful differences in how students acquire and apply programming patterns. Unlike traditional knowledge tracing over predefined skills, this approach tracks mastery of recurring structural patterns directly from student code, revealing whether students steadily improve, struggle intermittently, or persistently fail on specific programming patterns. These findings reinforce the validity of the KC model and motivate our ablation study presented in the following section.


\subsection{KC Model Comparison and Ablation Study}

To further evaluate the effectiveness of our proposed framework, we tested its ability to explain and predict student performance within the Additive Factors Model (AFM)~\cite{cen2006learning,koedinger2012automated}. The AFM estimates the probability of a correct response as a function of KC-specific intercepts and learning rates by fitting logistic learning curves. Following prior research~\cite{shi2023kc,alencar2025integrating}, we compared the performance of our KC model against two common baselines: the single-KC model, which treats all submissions as practicing the same KC, and the item-KC model, which assigns each problem its own KC. Model performance was evaluated using three metrics: Akaike Information Criterion (AIC), Bayesian Information Criterion (BIC), and Root Mean Squared Error (RMSE). Lower AIC and BIC indicate better fit, with BIC applying a stronger penalty for model complexity, while RMSE reflects average prediction error. 

We also conducted an ablation study to isolate the contribution of our combined approach, including the SANN and $\beta$-VAE pipeline. For this purpose, two additional baselines were tested: (i) replacing the VAE model by using embeddings from a large programming language model, jina-code-embeddings (1.5b parameters)~\cite{kryvosheieva2025efficient}, (ii) replacing the VAE model with a fine-tuned CodeT5 model (pretrained programming language model, 220 million parameters)~\cite{wang2021codet5}, using its encoder output in the clustering step and (iii) removing the VAE entirely and clustering directly on the SANN-generated embeddings. 

As shown in Table~\ref{tab:fit}, our pattern-based KC model outperformed all baselines. It achieved lower AIC and BIC values, indicating better model fit with appropriate complexity, and a lower RMSE, demonstrating improved predictive accuracy of student performance. These results suggest that our framework provides a more accurate and effective representation of KCs compared to traditional skill-attribution or direct embedding strategies. We next examine the explainability of pattern-based KCs, focusing on how the framework surfaces the code structures underlying each discovered KC.

\begin{table}
\centering
\caption{KC model comparison (↓ = lower is better).}
\begin{tabular}{|l|c|c|c|} \hline
\textbf{Model} & \textbf{AIC $\downarrow$} & \textbf{BIC $\downarrow$} & \textbf{RMSE $\downarrow$} \\ \hline
Single KC & 60,327.41 & 60,344.95 & 0.50 \\ \hline
Item KC & 57,496.26 & 58,372.99 & 0.48 \\ \hline
jina-code-embeddings & 60832.70 & 61704.36 & 0.48 \\ \hline
CodeT5-based KC & 57,300.44 & 58,155.97 & 0.47 \\ \hline
SANN-based KC & 47,941.18 & 48,737.84 & 0.40 \\ \hline
Pattern-based KC & \textbf{39,319.73} & \textbf{40,196.47} & \textbf{0.35} \\ \hline
\end{tabular}
\label{tab:fit}
\end{table}

\begin{figure*}[t]
  \centering
  \subcaptionbox{Attempt 1 (incorrect, sortaSum)\label{fig:sub1}}%
    {\includegraphics[height=0.21\textheight]{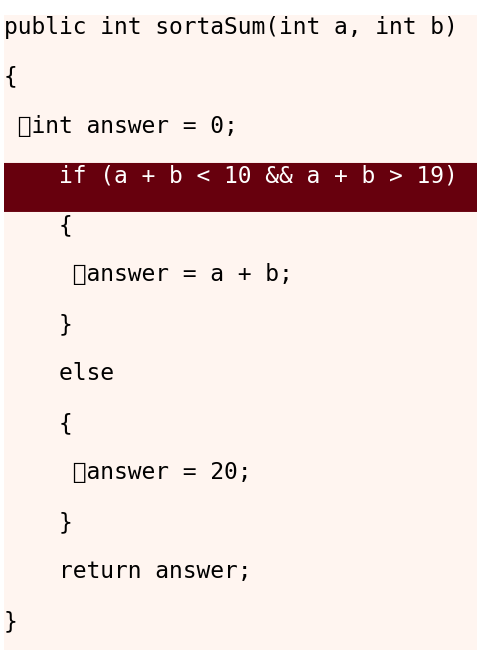}}\hfill
  \subcaptionbox{Attempt 2 (incorrect, sortaSum)\label{fig:sub2}}%
    {\includegraphics[height=0.21\textheight]{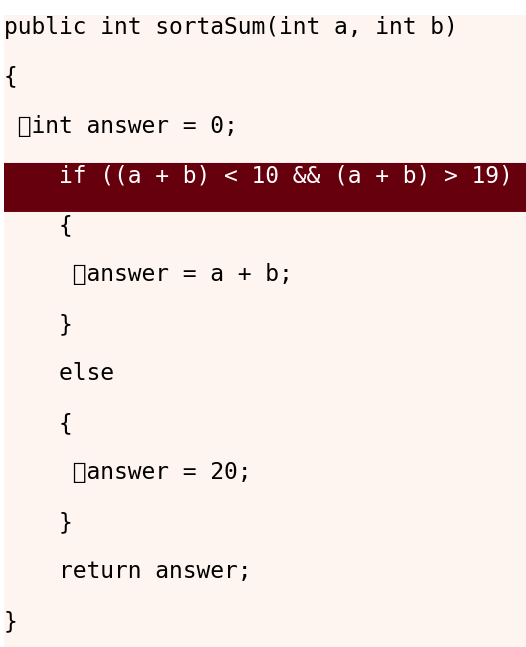}}\hfill
  \subcaptionbox{Attempt 3 (correct, sortaSum)\label{fig:sub3}}%
    {\includegraphics[height=0.21\textheight]{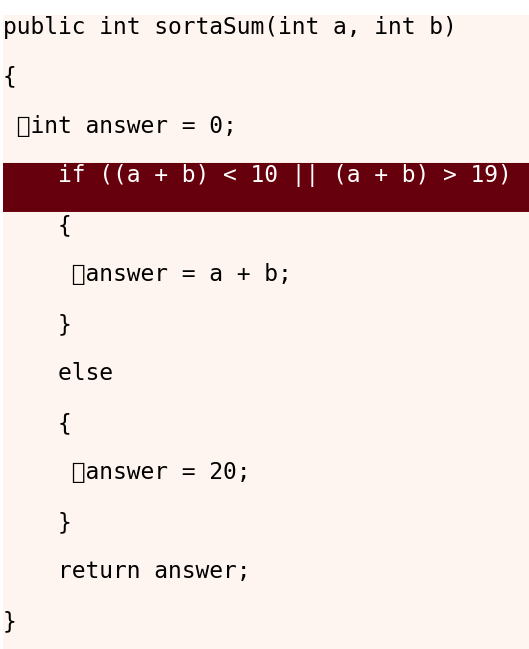}}\hfill
  \subcaptionbox{Correct, int1To10 \label{fig:sub4}}%
    {\includegraphics[height=0.21\textheight]{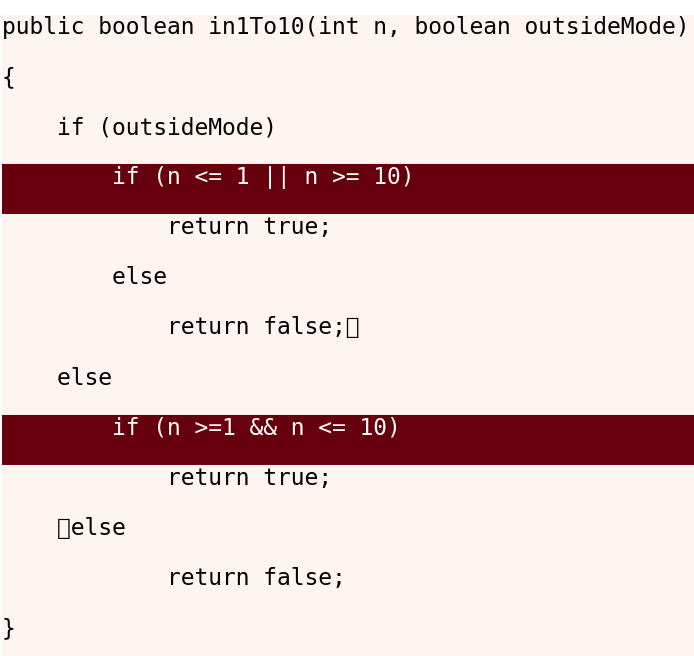}}

  \caption{KC~42 across problems. (a--c) show three sequential attempts on the \textit{sortaSum} problem, where KC~42 is incorrectly (a--b) and then correctly (c) implemented. 
  (d) shows a correct implementation of the same KC in the \textit{int1To10} problem.}
  \label{fig:kc42_explain}
\end{figure*}

\subsection{Pattern-based KC Cluster Evaluation}
While predictive performance and learning curve behavior provide indirect evidence of KC quality, it is also important to verify that the discovered clusters correspond to educationally meaningful and internally coherent programming patterns. To this end, as described in Section~\ref{method:educational_utility}, we conducted an expert evaluation of $35$ pattern-based KC clusters, which showed a clear learning trend. The expert evaluation results indicate that the discovered pattern-based KCs are both educationally meaningful and internally consistent. All evaluated cluster medoids ($100$\%) were judged to represent important programming patterns appropriate for introductory programming instruction. These medoids commonly captured reasoning patterns such as compound conditionals, guarded logic, accumulator updates, and loop–condition interactions, suggesting that the clustering process consistently surfaces instructionally relevant concepts rather than trivial or boilerplate code fragments.

Cluster consistency was also high. Across all clusters, $97.7$\% of candidate patterns were judged to express the same underlying programming idea as their corresponding medoid, despite syntactic variation and differences in problem context. In addition, candidate patterns showed strong similarity in how they manipulated program state or control flow, with an average action-level similarity score of $3.7$ (std $=$ $0.65$) on a $1$–$4$ scale. The high mean is expected, as candidate patterns were sampled from within-cluster neighborhoods around each medoid, and the goal of this evaluation was to assess cluster coherence. 

Altogether, the proposed framework yields stable pattern-based KCs. The resulting clusters align with important programming constructs and form coherent units for downstream analyses, including knowledge tracing, learning curve evaluation, and explainable feedback generation.


\subsection{Explainability of Pattern-based KCs}

A central goal of our framework is not only to model performance but also to make visible the reasoning behind predictions, toward actionable feedback. To do so, we project attention-weighted subtrees back onto source code and label each span with its associated pattern-based KC. This process reveals the syntactic constructs driving the model’s decisions, allowing for precise diagnosis of misconceptions.

Figure~\ref{fig:kc42_explain} illustrates this process using pattern-based KC~42, which governs the correct use of compound comparison and logic operators in conditional expressions. In a sequence of three attempts on the \textit{sortaSum} problem, a randomly selected student first mis-implements KC~42, misapplying \texttt{\&\&} instead of \texttt{||} (Figures~\ref{fig:sub1}, \ref{fig:sub2}); the model highlights the faulty condition in red. In the third attempt (Figure~\ref{fig:sub3}), the student corrects the logic, and the highlight persists, indicating the KC remains relevant while now being mastered. Figure~\ref{fig:sub4} shows another instance of KC~42 in a correct submission for the \textit{int1To10} problem, where students must reason about boundary conditions. Despite the different problem context, the conditional again reflects KC~42, validating its conceptual consistency across problems and representing how our framework discovers reusable conceptual patterns that extend beyond individual assignments.


These visualizations show how our framework links model predictions to code structures, making visible the constructs that drive performance outcomes. They also demonstrate the cross-problem consistency of pattern-based KCs, showing that the same conceptual pattern can be recognized across different assignments. In this way, the framework not only clarifies individual predictions but also represents reusable knowledge structures that make KC tracking more generalizable and pedagogically meaningful. Having established the validity and utility of pattern-based KCs, we discuss their implications for student modeling, programming education, and future research in the following section.



\section{Discussion}
This work set out to address the challenge of defining and modeling knowledge components in introductory programming, a domain where solution paths are highly variable and traditional skill lists are difficult to specify. We introduced \emph{pattern-based KCs}, defined as recurring structural patterns in student code that represent the programming constructs involved in solving problems.

To operationalize this definition, we developed a pipeline to discover pattern-based KCs from student code submissions. Using the SANN model, we identified AST substructures most relevant to code correctness. These substructures were then abstracted with a variational autoencoder (VAE) to capture structural similarities across submissions and clustered into representative KCs. The resulting set of pattern-based KCs advances student modeling by enabling knowledge tracing at a more fine-grained level of learner behavior. Our proposed KC-DKT model improved predictive performance by more than 13\% over the original DKT model, and learning curve analysis further confirmed that the discovered KCs align with established learning theory. In addition, an expert evaluation of the clusters found that all medoid patterns were judged educationally useful and that within-cluster samples showed high conceptual and action-level consistency, providing evidence that the extracted KCs correspond to coherent instructional units.

Our approach offers several advantages over traditional methods. The integration of SANN provides transparency by enabling visualization of attention weights on specific code structures, allowing educators to understand and inspect each KC. Moreover, each KC can be traced back to the original student code patterns, supporting explainable, transparent practices for AI in education. This stands in contrast with existing deep learning and LLM-enabled approaches for KC discovery, which often operate as black boxes for KC definition and attribution. Additionally, clustering of the VAE-generated AST substructures captures a wide range of programming patterns, from simple granular patterns to complex logic. This breadth reflects the diversity of student solution strategies and ensures that the resulting KCs capture the full spectrum of patterns students use in practice, enabling more precise modeling of student knowledge in CS education.

These findings have direct implications for adaptive learning systems. By accurately modeling student mastery at the KC level in submitted code, our approach facilitates real-time feedback on specific coding patterns, enabling personalized learning approaches in CS education~\cite{corbett1997intelligent}. For example, with the identification provided by our approach, if a student consistently struggles with a particular loop construct, an adaptive learning system can provide targeted exercises or resources associated with that pattern to address this gap~\cite{abdelrahman2023knowledge}. Pattern-based KCs can also inform content recommendation engines, guiding the selection of problems~\cite{baker2006adapting} or worked examples~\cite{hosseini2020improving} that align with their current knowledge and learning needs.

Beyond individual feedback, aggregated KC-level data can help instructors identify which patterns are most challenging across a cohort, informing curriculum adjustments and instructional focus. Over time, this integration of KC analytics into adaptive platforms can create a feedback loop between student performance data and instructional design, ensuring that learning environments adapt to address the most persistent knowledge gaps. In this way, our work connects automated KC discovery with practical instructional use, providing an explainable and scalable framework for modeling student learning in CS education.

\section{Limitations and Future Work}

The results demonstrate the effectiveness of discovering pattern-based KCs from AST structures with attention-based modeling, although there are limitations that can be addressed through future research. First, because the dataset does not include demographic or background information, we are unable to analyze potential differences in KC distributions or learning trajectories across student subgroups; examining fairness and generalizability across populations is an important direction for future work. Second, our method relies on successful AST parsing, which excludes uncompilable submissions ($15$\% of the dataset) and removes potentially valuable data related to misconceptions or syntax errors~\cite{shi2023kc,hoq2025automated}. Future work could incorporate error-tolerant parsers or automated syntax-error repair tools that can handle uncompilable code~\cite{pitts2025automated}.

Moreover, our evaluation followed standard practice in knowledge tracing by focusing on first submission attempts to avoid noise from feedback-driven debugging. While this reduces bias, it also limits the ability to study repeated practice effects for KCs. Modeling post-first-attempt behavior, and distinguishing genuine skill development from surface-level fixes~\cite{baker2011gaming}, is an important next step. Another limitation of the current work is that all code representations are learned solely from the CodeWorkout dataset, without leveraging large pretrained code models or large language models (LLMs) to warm-start token or node embeddings. While this design choice allows the discovered pattern-based KCs to remain tightly grounded in course-specific student behavior, it also constrains the representational capacity of the model, particularly given the relatively modest size of the dataset. Pretrained code embeddings have been shown to capture broad syntactic and semantic regularities across diverse programming contexts, which could improve generalization and stability of learned representations. An important direction for future work is therefore to initialize node- or subtree-level embeddings using pretrained code representations (e.g., from LLMs or large-scale code models), followed by fine-tuning on course-specific student data. More broadly, the framework could benefit from a two-stage training strategy in which models are pretrained on large, heterogeneous programming corpora and then adapted to individual courses or assignments, enabling the discovery of both general programming patterns and course-specific KCs. Hybrid pretraining–fine-tuning approaches may further improve the robustness, scalability, and transferability of pattern-based KC discovery.


Another important direction for future work should be the generation of natural language descriptions for the discovered pattern-based KCs. Currently, KC representations exist as structural subtrees, which may not be easily interpretable by novice learners or instructors. Using the natural language generation capabilities of large language models, future work could generate human-readable explanations or summaries of each KC. Further, integrating these KCs into a recommendation system for CS1 courses that provides targeted practice problems and worked examples could enable our framework to model student knowledge and actively support learning in real time. This would further enhance the accessibility, pedagogical utility, and feedback generation potential of our framework in educational settings.

\section{Conclusion}

In this work, we presented a framework for discovering pattern-based KCs from student programming submissions by combining the SANN model, a variational autoencoder, and clustering. We evaluated the resulting pattern-based KCs through integration with DKT, which produced significant improvements in predictive performance, and through learning curve analysis, which confirmed their alignment with learning theory. Our approach advances knowledge modeling in CS education by providing scalable, structurally grounded representations of student understanding. Beyond predictive modeling, these KCs also create opportunities for targeted practice, adaptive feedback, and personalized instructional support in programming education.

\section{Acknowledgments}
This research was supported by the National Science Foundation (NSF) under Grants \#2418658, \#2418655, and \#2418657. Any opinions, findings, and conclusions expressed in this material are those of the authors and do not necessarily reflect the views of NSF.

%
\bibliographystyle{abbrv}
\bibliography{sigproc}  
%
\end{document}